\documentclass[letterpaper, 10 pt, conference]{ieeeconf}
\IEEEoverridecommandlockouts
\let\labelindent\relax
\usepackage{enumitem}
\usepackage{balance}
\usepackage[font={small}]{caption}
\usepackage[labelformat=simple]{subfig}
\usepackage{array}
\usepackage{textcomp}
\usepackage{mathtools, nccmath}
\usepackage{graphicx}
\usepackage{amsfonts}
\usepackage{amsmath}
\usepackage{autobreak}
\usepackage{amssymb}
\usepackage{algorithm}
\usepackage{algorithmic}
\usepackage{tikz}
\usepackage{arydshln}
\usepackage{multirow}
\usepackage{bm}
\usepackage{epstopdf}
\usepackage{cite}
\usepackage{siunitx}
\usepackage{xcolor}
\usepackage{float} 
\usepackage{stfloats}
\usepackage{lineno}
\usepackage{subfloat}
\usepackage{colortbl}
\usepackage{booktabs}

\makeatletter
\let\NAT@parse\undefined
\makeatother
\usepackage[colorlinks=true,
            linkcolor=blue,
           ]{hyperref}

\allowdisplaybreaks[4]

\title{\LARGE \bf Quadruped Guidance Robot for the Visually Impaired: \\
A Comfort-Based Approach}
\author{Yanbo Chen$^\dagger$, Zhengzhe Xu$^\dagger$, Zhuozhu Jian$^\dagger$, Gengpan Tang, Liyunong Yang, \\Anxing Xiao, Xueqian Wang, Bin Liang
\thanks{$^\dagger$ indicates equal contribution.}
\thanks{
This work was supported by the National Key R\&D Program of China 2022YFB4701400/4701402 and the National Natural Science Foundation of China No.62293545. Corresponding authors: Anxing Xiao, Xueqian Wang.
}
\thanks{Yanbo Chen, Zhengzhe Xu, Gengpan Tang, and Liyunong Yang are with the School of Mechanical Engineering and Automation at Harbin Institute of Technology, Shenzhen 518055, China, \tt\{190320129, 200320314,190310113,190320329\}@stu.hit.edu.cn}
\thanks{Zhuozhu Jian, Xueqian Wang, and Bin Liang are with the Center for Artificial Intelligence and Robotics, Shenzhen International Graduate School, Tsinghua University, Shenzhen 518055, China, \tt \{jzz21@mails., wang.xq@sz., liangbin@\}tsinghua.edu.cn}
\thanks{Anxing~Xiao is with the School of Computing, National University of Singapore, Singapore 117417, Singapore. {\tt anxingx@comp.nus.edu.sg} }
}

\begin{document}

\maketitle
\begin{abstract}

Guidance robots that can guide people and avoid various obstacles, could potentially be owned by more visually impaired people at a fairly low cost. 
Most of the previous guidance robots for the visually impaired ignored the human response behavior and comfort, treating the human as an appendage dragged by the robot, which can lead to imprecise guidance of the human and sudden changes in the traction force experienced by the human. 
In this paper, we propose a novel quadruped guidance robot system with a comfort-based concept. We design a controllable traction device that can adjust the length and force between human and robot to ensure comfort.
To allow the human to be guided safely and comfortably to the target position in complex environments, our proposed human motion planner can plan the traction force with the force-based human motion model. 
To track the planned force, we also propose a robot motion planner that can generate the specific robot motion command and design the force control device. 
Our system has been deployed on Unitree Laikago quadrupedal platform and validated in real-world scenarios.
(Video\footnote{Video demonstration: \url{https://youtu.be/gd-RcYOqGuo}.})

\end{abstract}

\section{Introduction}
\label{sec:Introduction}

Guide dogs can lead the visually impaired to avoid obstacles, assist them with their daily activities and improve their quality of life. 
However, the number of guide dogs is extremely limited due to the high cost of time and money in training. 
In recent years, quadruped robots have displayed the ability to achieve challenging dynamic motions, such as jumping over obstacles \cite{gilroy2021autonomous} and navigating through uneven terrain \cite{mastalli2020motion}.
Compared to high-cost guide dogs, quadruped robots are likely to be owned by more visually impaired people because of their high reproducibility and agility.

In previous guidance robot studies, whether the human and the robot were linked by a rigid arm \cite{borenstein1997guidecane,chuang2018deep,li2019toward,hamed2019hierarchical} or an inelastic leash \cite{xiao2021robotic}, they did not take into account the human comfort, leading to a wide variation in traction force.
A recent study \cite{rubagotti2022perceived} discusses perceived safety (including comfort) affecting physical human-robot interaction (pHRI), noting that significant factors include robot speed, motion fluency and predictability, and smooth contacts. 
Our goal in this paper is to propose a quadruped guidance robot system that can safely guide the human in a narrow indoor environment while taking comfort into account. 
Human motion depends on the traction force and the human’s current state. When the traction force is appropriate, human comfort will be significantly improved. Therefore, we seek to emphasize the force in the traction device and to design the corresponding comfort-based planning and control system.

\begin{figure}
    \centering
    \includegraphics[width=8.5cm]{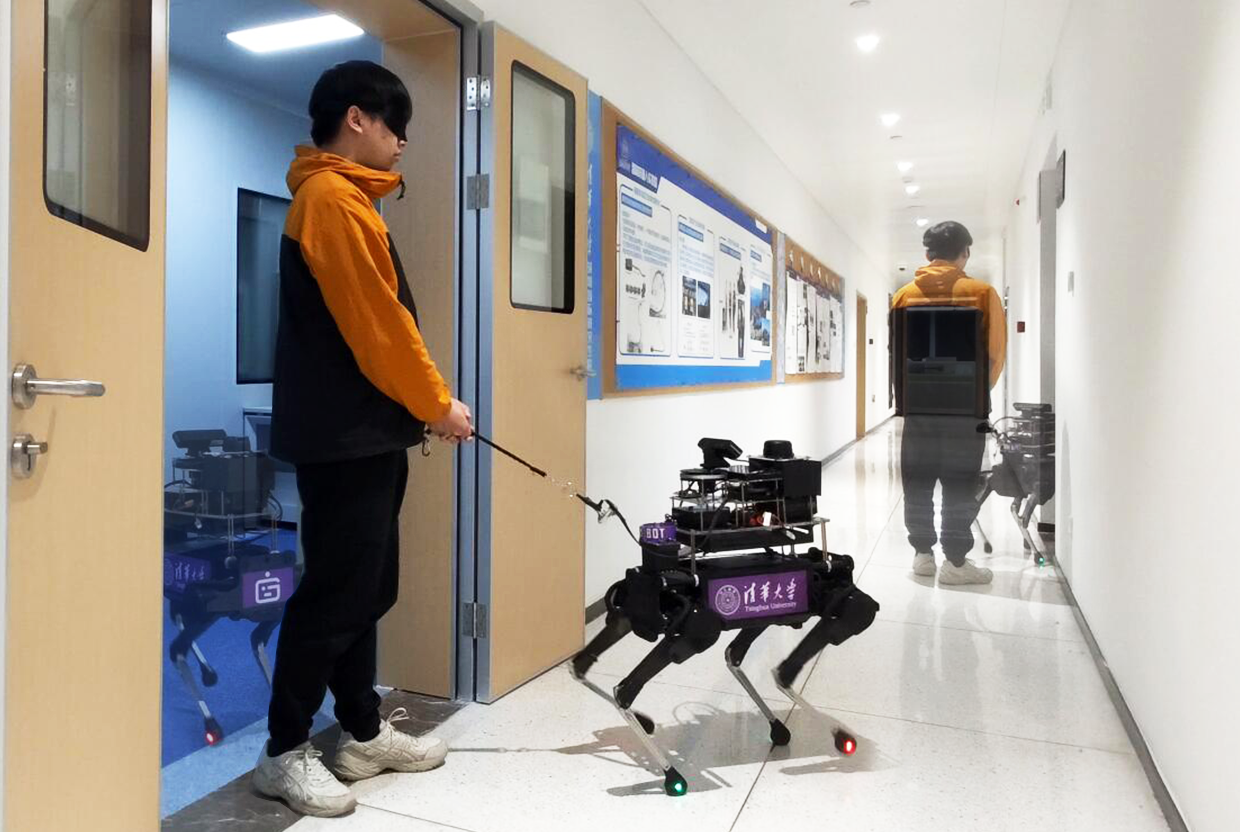}
    \caption{
    A blind-folded person is guided by Laikago to turn out and cross the corridor. The person is guided by an elastic rope and a force control device is used to control the force to ensure comfort.
    }
    \label{fig:main}
    \vspace{-0.2cm}
\end{figure}

\subsection{Related Work}
\label{sub:related}
Guidance assistive technologies capable of guiding the visually impaired are a long-term study. 
As early as 1980, the guide dog robot MELDOG \cite{tachi1981guide} successfully achieved the guiding task through experimental hardware. 
Initially, guidance robots mostly used wheeled platforms \cite{hsu2009hybrid,soltani2014pointwise,guerreiro2019cabot,bruno2019development}, which had limitations in complex terrain. 
With the advancement of robotic mobile platforms, legged robots \cite{hamed2019hierarchical,xiao2021robotic} and drones \cite{folmer2015exploring,Tan2021FlyingGD} are also used for guidance, exerting superior motion performance. 
Despite this, drones have insufficient load capacity for installing sensors, whereas legged robots are more viable due to their strong load capacity and agility.

Guiding the human is a complicated pHRI task. 
Modalities include haptic feedback-based guidance \cite{katzschmann2018safe} and auditory guidance \cite{kalpana2020voice,yang2020indoor}, as well as traction.
However, haptic feedback approaches may not guarantee comfort due to vibration, and auditory-based methods may be susceptible to noise disturbance and may cause uncertainty in movement. 
Currently, most studies are based on traction guidance, including the use of a rigid connection \cite{borenstein1997guidecane,chuang2018deep,ye2016co} or a flexible connection\cite{tachi1981guide,folmer2015exploring,xiao2021robotic} between a human and a robot. 
Rigid connections may result in large changes in the traction force on a human when walking or stopping suddenly.
Using a leash connection is more flexible, and \cite{xiao2021robotic} considers a hybrid system model that includes leash slack and tightness, but the long planning time may cause anxiety, as well as discomfort when the leash changes repeatedly from slack to taut. 
Another critical aspect that necessitates attention is the absence of proper planning and control of traction force in previous research.
Although \cite{xiao2021robotic} proposed the force of constraint to enable the planning behavior to be more comfortable, it ignores the real-time feedback control of the traction force.

To address the above problems, we aim to establish a guidance robot system with a comfort-based concept through force interaction. To solve appropriate traction forces (magnitude and direction), we develop a force-based human motion model considering human ``standing-walking'' motion pattern with reference to the minimum-jerk theory \cite{1987The} and apply it in the human motion planner. A controllable traction device is introduced to enable force control. Given the expected forces, the robot motion planner generates the velocity commands for the robot to physically guide the human. The overall comfort-based system enables the visually impaired to traverse cluttered spaces while reducing discomfort. 

\subsection{Contributions}

    This work offers the following contributions:
    \begin{itemize}
        \item A novel autonomous guidance robotic system is introduced, featuring a controllable traction device and a planning and control framework based on comfort. This system enables precise control of traction force and promotes smooth interaction.
        
        \item A force-based human motion model is developed to describe the human “standing-walking” motion pattern in a robotic guidance system, enabling traction force planning. The model is validated with experimental data and the parameter is adaptively estimated.
        
        \item A human motion planner and a robot motion planner are proposed to achieve collision-free human guidance during navigation to the goal location while considering the planning and control of traction forces.

        \item The guiding system is deployed on the Unitree Laikago quadruped platform and validated in the experiment with real-world scenarios. Comparative experiments quantitatively and qualitatively demonstrate that the proposed system can significantly improve the comfort of guidance. 
    \end{itemize}
\section{System Design}
\label{sec:framework}
\subsection{Hardware System}
\label{subsec:mechanical}

\begin{figure}[!htbp]
    \centering
    \includegraphics[width=8.65cm]{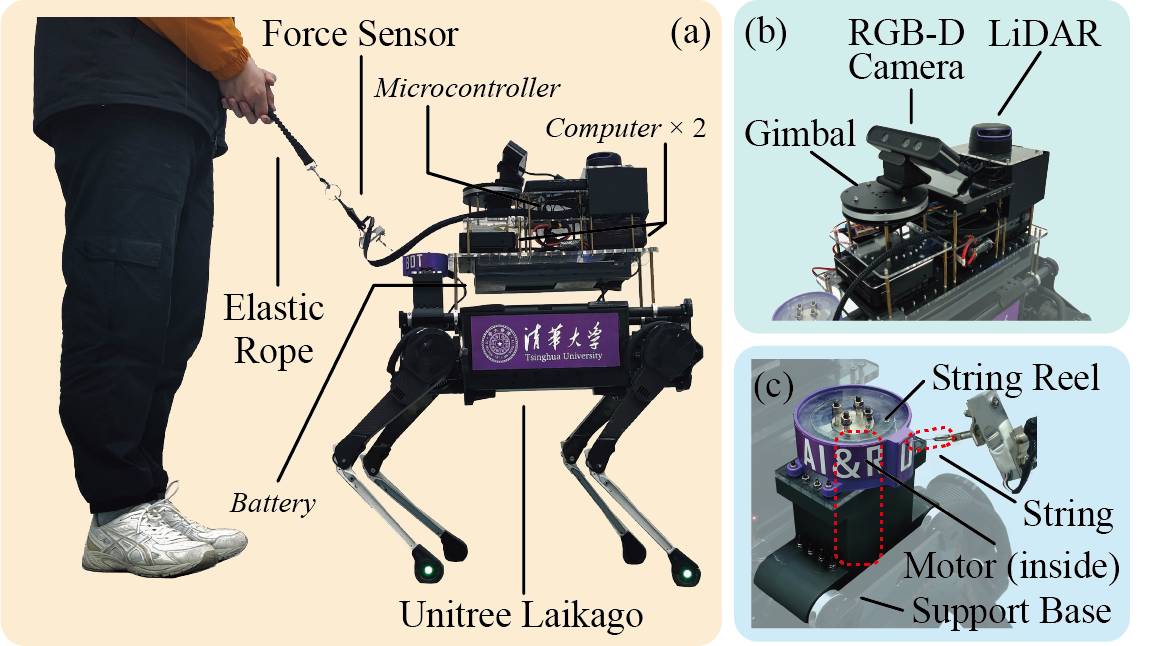}
    \caption{
    The guidance robot consists of the quadruped platform (Fig. \ref{fig:robotic guide dog}(a)), the sensors module (Fig. \ref{fig:robotic guide dog}(b)), and the force control device module (Fig. \ref{fig:robotic guide dog}(c)). Two high-performance computers and a high-capacity battery are equipped to support stable operation (Fig. \ref{fig:robotic guide dog}(a)).
    }
    \label{fig:robotic guide dog}
    \vspace{-0.3cm}
\end{figure}
The quadruped guidance robot, depicted in Fig. \ref{fig:robotic guide dog}, comprises three main components:

\emph{1) Robotic Platform}: Unitree Laikago is a quadruped robotic platform, with a length of 0.65m, a width of 0.35m, a height of 0.6m, a maximum load of 5kg and a walking speed of -0.5$\sim$0.8m/s. 
The platform's robust load capacity enables more equipment and complex mechanisms to be integrated into the system.

\emph{2) Sensors}: A 2D LiDAR is used for perception and localization. 
    An RGB-D camera is mounted on the 1-DoF gimbal to detect the human's position. 

\begin{figure}[!htbp]
   \vspace{-0.3cm}
    \centering
    \includegraphics[width=8.8cm]{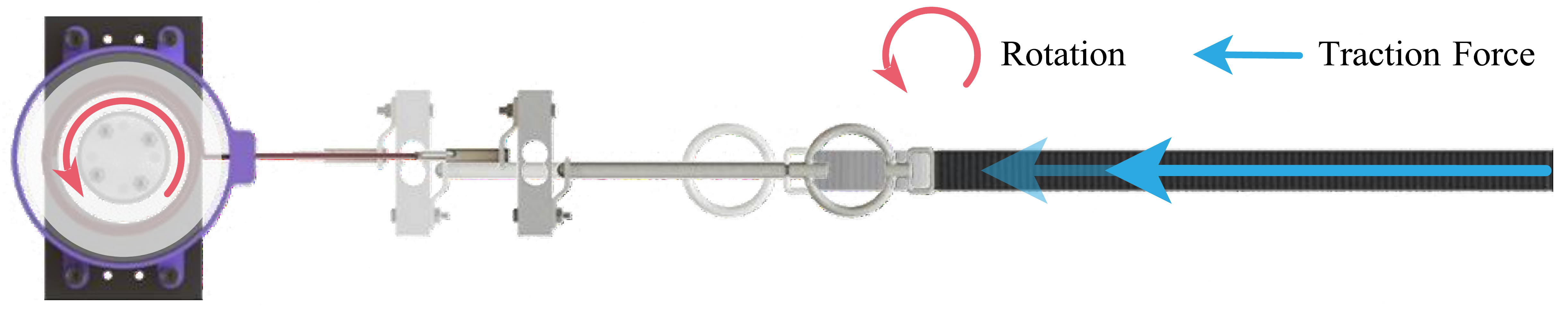}
    \caption{
    Illustration of the traction device (top view). The motor rotates and retracts the string, causing the elastic rope longer and the traction force larger.
    }
    \label{fig:TD}
    \vspace{-0.5cm}
\end{figure}

\emph{3) Traction Device}: The traction device is mounted at the rear of the robot. It consists of an elastic rope and a force control device (FCD) containing a support base, DC gear motor, acrylic string reel, thin string, force sensor, motor PID controller Arduino MEGA2560.
    Based on Hooke's Law $F = K\Delta l$, where $K$ is the stiffness coefficient and $\Delta l$ is the elongation, the magnitude of the traction force depends on the length of the elastic rope. 
    As demonstrated in Fig. \ref{fig:TD}, when the distance between the human and the robot remains constant, 
    the overall length of the elastic rope and the thin string is fixed.
    The motor rotation can retract or release the string, thus adjusting the elastic rope's length.
    Since the speed of retracting and releasing the string is much higher than the relative speed between human and robot, it can be assumed that the change of elastic rope's length in any short time depends on the motor speed, that is, motor speed determines the change in force magnitude. 
    The motor controller receives the desired force magnitude from the computer and the feedback from the force sensor, calculates the required motor speed, and controls the motor, thereby controlling the length of the elastic rope and finally achieving the desired force.

\subsection{System Framework}
\label{subsec:framework}
\begin{figure*}
    \centering
    \includegraphics[width=17.6cm]{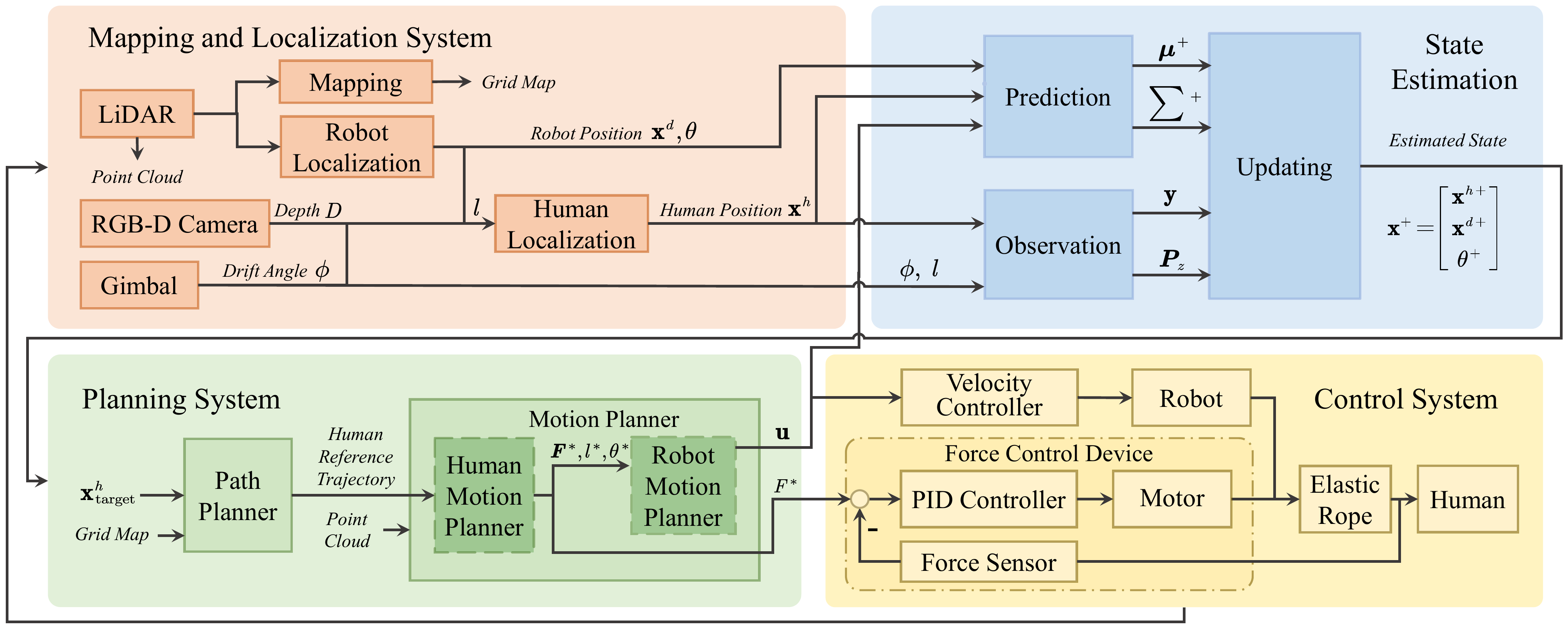}
    \caption{
    Overview of system framework. The mapping and localization system obtains the system state. After state estimation, the estimated state is passed to the planner. The planner uses the grid map and estimated state to solve for the reference force and control inputs.
    }
    \label{fig_framework}
    \vspace{-0.5cm}
\end{figure*}
    As shown in Fig. \ref{fig_framework}, the guiding system can be divided into four modules: 
    mapping and localization system, state estimation, planning system and control system. 
    
    In the mapping and localization system, the LiDAR is used for mapping and robot localization. 
    LiDAR, gimbal, and RGB-D camera jointly measure the human position. 
    To reduce the error, the Unscented Kalman Filter is used for state estimation, as detailed in Sec.\ref{subsec:UKF}. 
    
    During guidance, the planning system utilizes the grid map, point cloud and estimated state to perform path planning and motion planning successively, while continuously updating and replanning. 
    The path planner follows the approach described in Sec.\ref{subsec:Path Planning}
    to obtain a collision-free human path and passes it to the motion planner. 

    Considering that the human step frequency is relatively low ($<$5Hz) while the robot and motor can be both controlled at a high frequency, we separate human motion planning from robot motion planning.
    The desired forces (magnitude and direction) are computed in the human motion planning described in Sec.\ref{subsec:Motion Planning}. 
    Concurrently, the robot motion planning is performed in a separate thread based on MPC to control the force direction, while the motor controller performs high-frequency PID control of the force magnitude. 
    Finally, the human moves along the desired trajectory under the traction of the elastic rope controlled by the robot and the force control device. 
\section{Methods} \label{sec:methods}
\subsection{System Formulation}
\label{subsec:formulation}
\begin{figure}[!htbp]
    \centering
    \includegraphics[width=7cm]{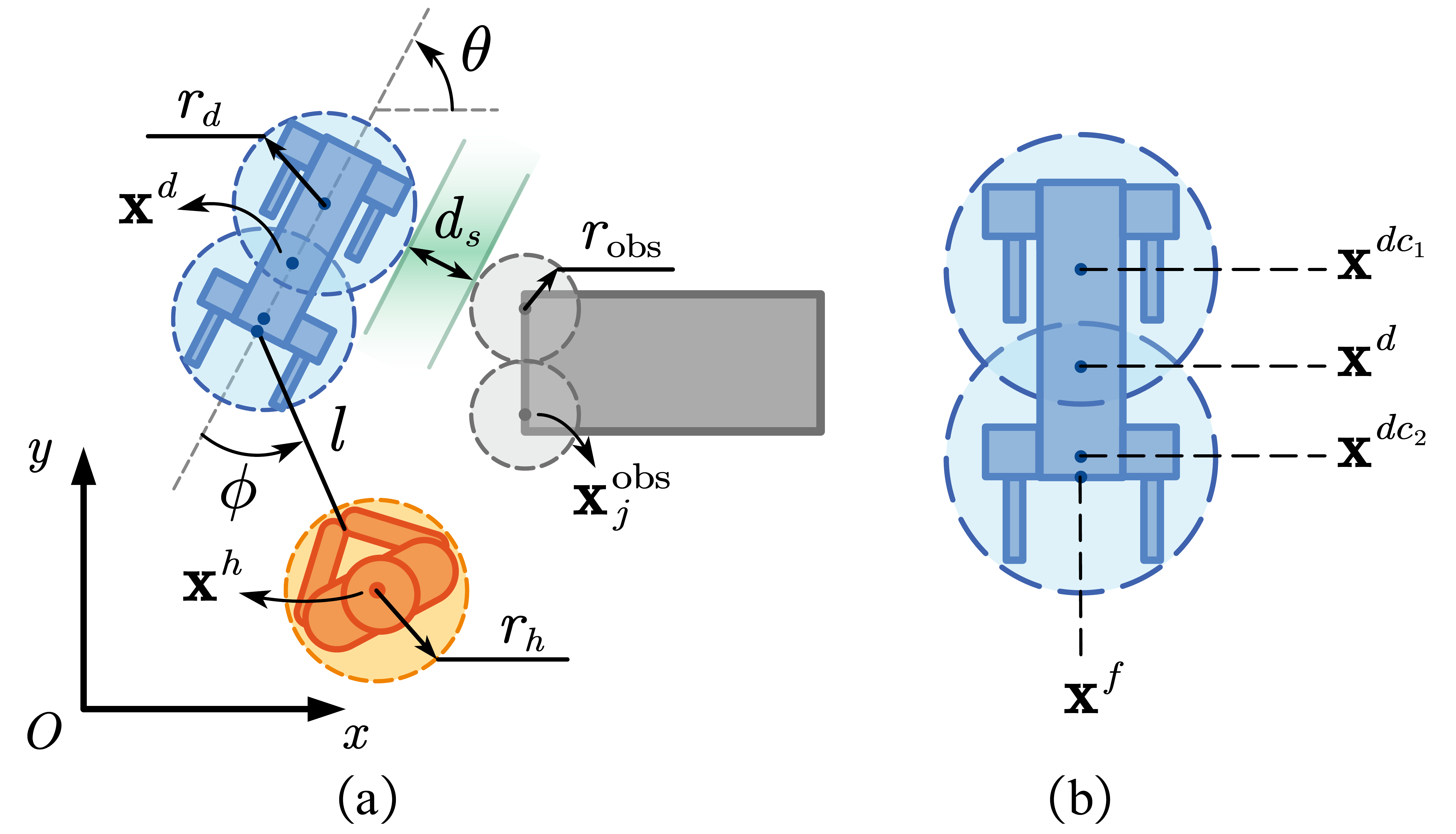}
    \caption{
    Configuration of the guiding system. Human is guided by the traction device. System state is  described by $\left( x^h,y^h,x^d,y^d,\theta \right)$ (Fig. \ref{fig:system formulation}(a)). Four key points $\mathbf{x}^d$, $\mathbf{x}^f$, $\mathbf{x}^{dc_1}$, $\mathbf{x}^{dc_2}$ are set on the robot (Fig. \ref{fig:system formulation}(b)).
    }
    \label{fig:system formulation}
    \vspace{-0.5cm}
\end{figure}

The guiding system consists of a human and a quadruped robot connected by the traction device, as shown in Fig. \ref{fig:system formulation}(a). 
To facilitate the analysis, we model the system in the 2D plane. 
We define system state as 
$
\mathbf{x}=\left( \mathbf{x}^{h\top},\mathbf{x}^{d\top},\theta \right) ^{\top}\in \mathbb{R} ^5
$ in the world frame, 
where 
$\mathbf{x}^h=\left( x^h,y^h \right) ^{\top}$ and $\mathbf{x}^d=\left( x^d,y^d \right) ^{\top}$
are the positions of human and robot respectively, 
and $\theta$ is the yaw angle of the robot. 
In addition, several key points are set on the robot.
As shown in Fig. \ref{fig:system formulation}(b), $\mathbf{x}^f=\mathbf{x}^d-d_{\mathrm{cf}}\mathbf{e}_{\theta}(\mathbf{e}_{\theta}=\left( \cos \theta ,\sin \theta \right) ^{\top})$ is the fixed point of the traction device, where $d_{\mathrm{cf}}$ is the distance between the fixed point and the robot center. 
Furthermore, $\mathbf{x}^{dc_1}$, $\mathbf{x}^{dc_2}$ are the centers of the expansion circles for obstacle avoidance, and the vector from the human to the rear of the robot is defined as  $\boldsymbol{l}=\mathbf {x}^f-\mathbf{x}^h=l\mathbf{e}_l$ ($\mathbf{e}_a$ for the unit direction vector of $\boldsymbol{a}$). 
\vspace{-0.65cm}

 \subsection{Force-Based Human Motion Model}
\label{subsec:force-based state machine model}
Previous study \cite{xiao2021robotic} did not quantitatively consider the forces in the model because the forces could not be controlled in real time.
Benefiting from the robot's agile movement and the controllable traction device, we give priority to ensuring a more comfortable force when guiding. Therefore, we do not pay attention to the robot kinematics first, but consider the movement trend of a human under a certain magnitude and direction of force during walking. 
The force $\boldsymbol{F}=F\mathbf{e}_F \in \mathbb{R}^2$ acting on the person is the focus of the analysis of human motion. 
Obviously, the force is in the same direction as the elastic rope, i.e., 
$\mathbf {e}_F=\mathbf{e}_l$.

The human gait is discrete and can be represented by
$
    \mathbf{x}_{k+1}^{h}=\mathbf{x}_{k}^{h}+\boldsymbol{s}_k
$
, where $\boldsymbol{s}_k$ denotes one step. 
We determine the human velocity by taking the ratio of the average step size to the time interval, resulting in $\mathbf{v}^h=\boldsymbol{\bar{s}}/T $.
Thus, the discrete-time state equation of human can be written as
$
    \mathbf{x}_{k+1}^{h}=\mathbf{x}_{k}^{h}+\mathbf{v}_{k}^{h}T. 
$ When guided, the human's motion can be categorized into two states: standing and walking, denoted as $q_s$ and $q_w$, respectively. 

\emph{1) Standing State}: 
When the human is standing still, we have $\mathbf{v}^h=\mathbf{0}$. The robot position can be computed by
\begin{equation}
    \mathbf{x}^d=\boldsymbol{f}_d\left( \mathbf{x}^h,\boldsymbol{F},l,\theta \right) =\mathbf{x}^h+l\mathbf{e}_F+d_{\mathrm{cf}}\mathbf{e}_{\theta}, 
    \label{equ:dog_position}
\end{equation}
as shown in Fig. \ref{fig:system formulation}

\emph{2) Walking State}: 
When the human is walking, their movement direction aligns with the force direction, and the step size depends on the force magnitude. The experiment in Sec.\ref{subsec:system_identification} illustrates the relationship between the velocity of the human moving along the rope direction and the force as $\mathrm{v}^{hl}=\mathbf{v}^h\cdot \mathbf{e}_l=\alpha F+\beta (\alpha, \beta\in\mathbb{R})$. 
The robot's position can still be computed by \eqref{equ:dog_position}.

\begin{figure}[!htbp]
    \vspace{-0.3cm}
    \centering
    \includegraphics[width=8.5cm]{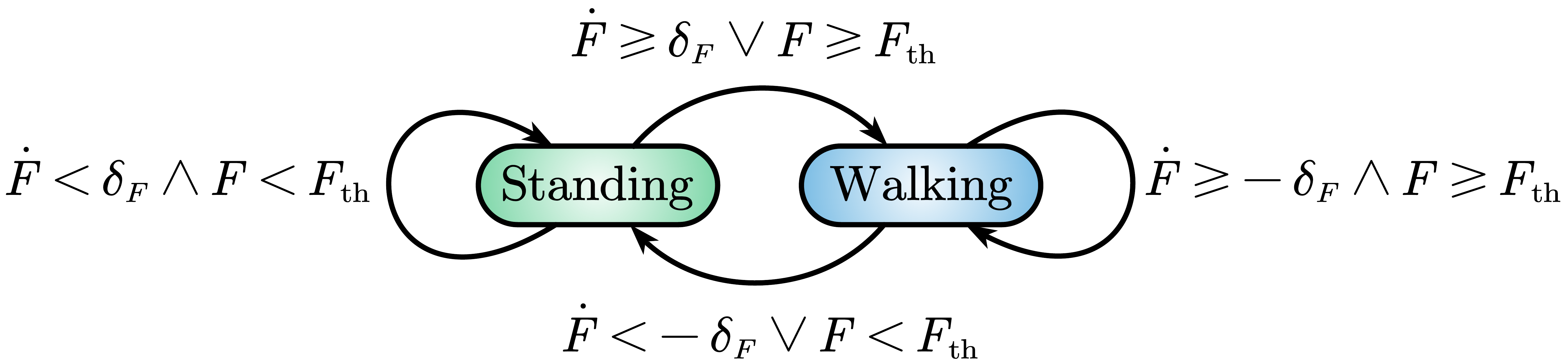}
    \caption{
    The human standing and walking state transition diagram, the transition is determined by the human's current state and force.
    }
    \label{fig:switch mode}
    \vspace{-0.3cm}
\end{figure}

\emph{3) State Transfer Condition}: 
A human switches from standing to walking depending on the rate of change of the applied force. According to the minimum-jerk theory \cite{1987The}, when a sudden change in force is acted on to the human's arm, human tends to minimize jerk (derivative of acceleration). Therefore, the human starts to walk in the force direction. 
To keep walking, the magnitude of the force needs to be greater than a certain threshold $F_{\mathrm{th}}$, otherwise the human will stop. 
Fig. \ref{fig:switch mode} depicts the above state transitions. 

We define $q_s=0$, $q_w=1$, and change in force in time interval $T$ as $\Delta F_k=F_{k+1}-F_k$. 
The discretized state transition equation is
\vspace{-0.2cm}
\begin{align}
    \label{equ:qk+1}
    q_{k+1}&=\delta \left( q_k,F_{k+1},F_k \right) \notag
    \\
    &=\begin{cases}
    	\begin{array}{c}
    	0,\mathrm{if}\left( q_k=0\land \left( \Delta F_k<\delta _F\cdot T\land F_k<F_{\mathrm{th}} \right) \right)\\
    	\lor \left( q_k=1\land \left( \Delta F_k<-\delta _F\cdot T\lor F_k<F_{\mathrm{th}} \right) \right)\\
    \end{array}\\
    	\begin{array}{c}
    	1,\mathrm{if}\left( q_k=0\land \left( \Delta F_k\geqslant \delta _F\cdot T\lor F_k\geqslant F_{\mathrm{th}} \right) \right)\\
    	\lor \left( q_k=1\land \left( \Delta F_k\geqslant -\delta _F\cdot T\land F_k\geqslant F_{\mathrm{th}} \right) \right)\\
    \end{array}\\
    \end{cases}.
    \vspace{-0.4cm}
\end{align}

Thus, the human state equation is given by
\vspace{-0.1cm}
\begin{align}
    \label{equ:xhk+1}
    \mathbf{x}_{k+1}^{h}=\boldsymbol{f}_h\left( \mathbf{x}_{k}^{h},\boldsymbol{F}_k,q_k \right) =\mathbf{x}_{k}^{h}+q_k\mathrm{v}_{k}^{hl}\mathbf{e}_{F_k}T.
    \vspace{-0.2cm}
\end{align}

\emph{4) Adaptive Parameter Estimation:} 
We design an adaptive identification module to obtain the parameters in the $F-\mathrm{v}^{hl}$ relationship of the first-time user. 
The module allows the robot to perform uniformly accelerated linear motion with force control device disabled and guides the human. 
By measuring the actual velocity and force, after low-pass filtering to reduce noise, the least-squares estimation is performed to obtain the model parameters.

\subsection{Motion Planning}
\label{subsec:Motion Planning}
Motion planning is essential for ensuring the safety and comfort of human locomotion. 
Our guiding system employs a two-stage approach to plan both human and robot motions.
In the first stage, the human motion planner will be performed using the force-based human motion model introduced in Sec.\ref{subsec:force-based state machine model} to obtain reference forces and pass it to the force control device. 
In addition, the optimized values from the first stage are passed to the robot motion planner to solve the robot control inputs by applying MPC-based trajectory tracking. 

\emph{1) Human Motion Planning}: 
The force-based human motion model is used to plan for the magnitude and direction of the force that can guide the human comfortably along the reference path.
The robot position constraint needs to be considered in the planning because the robot position determines the force direction and the robot obstacle avoidance is a fairly strong constraint. 
We achieve force planning optimal control by formulating the following MPC problem with an $N$-step horizon: 
\vspace{-0.3cm}
    \begin{subequations}
        \begin{align}
            \underset{\left\{ \boldsymbol{F}_k,l_k,\theta_k \right\}}{\min}
            &\left\| \mathbf{x}_{N}^{h}-\mathbf{x}_{\mathrm{target}}^{h} \right\| _{\mathbf{Q}_{t}^{h}}^{2}+\sum_{k=0}^{N-1}{(}\left\| \mathbf{x}_{k}^{h}-\mathbf{x}_{k}^{h*} \right\| _{\mathbf{Q}^h}^{2} \notag\\
            +&\left\| \boldsymbol{F}_{k+1}-\boldsymbol{F}_k \right\| _{\mathbf{Q}_F}^{2}+ w_{\Delta\theta}\left( 1-\cos \left( \theta _{k+1}-\theta _k \right) \right)  \notag\\
            +&w_l\left( K\left( l_{k+1}-l_k \right) -\left( F_{k+1}-F_k \right) \right)^2 ) \\
            \mathrm{s}.\mathrm{t}.\quad &\mathbf{x}_{0}=\mathbf{x}_{\mathrm{curr}},\quad q_{0}=q_{\mathrm{curr}}
            \label{current_constrant}
            \\
            &\mathbf{x}_{k+1}^{h}=\boldsymbol{f}_h\left( \mathbf{x}_{k}^{h},\boldsymbol{F}_k,q_k \right) 
            \label{human state transition}
            \\
            &\mathbf{x}_{k+1}^{d}=\boldsymbol{f}_d\left( \mathbf{x}_{k+1}^{h},\boldsymbol{F}_k,l_k,\theta _k \right) 
            \label{dog state transition}
            \\
            &q_{k+1}=\delta \left( q_k,F_{k+1},F_{k} \right) 
            \\
            &F_{\min}\leqslant \left\| \boldsymbol{F}_k \right\| \leqslant F_{\max}
            \label{F}
            \\
            &l_{\min}\leqslant l_k\leqslant l_{\max}
            \label{l}
            \\
            &\left< \mathbf{e}_{F_{k+1}},\mathbf{e}_{F_k} \right> \geqslant \cos \left( \varphi _F \right) 
            \label{F direction}
            \\
            &\left< \mathbf{e}_{F_k},\mathbf{e}_{\theta _k} \right> \geqslant \cos \left( \varphi _{\theta} \right) 
            \label{F-yaw}
            \\
            &\left\| \mathbf{x}_{k}^{h}-\mathbf{x}_{j}^{\mathrm{obs}} \right\| \geqslant d_s+r_h+r_{\mathrm{obs}}
            \label{obs_h_1}
            \\
            &\left\| \mathbf{x}_{k}^{dc_i}-\mathbf{x}_{j}^{\mathrm{obs}} \right\| \geqslant d_s+r_d+r_{\mathrm{obs}}\left( i=1,2 \right)
            \label{obs_d_1}
        \end{align}
    \end{subequations}
    where $\left\| \mathbf{x} \right\| _{\mathbf{Q}}\coloneqq \sqrt{\mathbf{x}^{\top}\mathbf{Qx}}$.
    $\mathbf{Q}_{t}^{h}$, $\mathbf{Q}^h$, $\mathbf{Q}_F \in \mathbb{R} ^{2\times 2}$ are positive definite, 
    $w_l$, $w_{\Delta\theta}$, $w_{\theta _0}\in \mathbb{R}$ are weights. $\left\{ \mathbf{x}_{k}^{h*} \right\} _{k=0}^{N}$ is the reference human path, $d_s$ is the safety margin, $r_h, r_d,r_\mathrm{obs} $ are the expansion radii of human, robot and obstacle, respectively. 
    $\varphi _F$ is the upper bound of the force direction angle change, and $\varphi _\theta$ is the
    upper bound of the difference between the yaw angle and the force direction angle.
    
    In the cost function, $\left\| \mathbf{x}_{k}^{h}-\mathbf{x}_{k}^{h*} \right\| _{\mathbf{Q}^h}^{2}$ tracks the reference path and $\left\| \mathbf{x}_{N}^{h}-\mathbf{x}_{\mathrm{target}}^{h} \right\| _{\mathbf{Q}_{t}^{h}}^{2}$ is the terminal cost. 
    The term $\left\| \boldsymbol{F}_{k+1}-\boldsymbol{F}_k \right\| _{\mathbf{Q}_F}^{2}$ smooths the force, and $w_{\Delta\theta}\left( 1-\cos \left( \theta _{k+1}-\theta _k \right) \right)$ minimizes the changes in yaw angle, which implies kinematic constraints.
    Further more, $w_l\left( K\left( l_{k+1}-l_k \right) -\left( F_{k+1}-F_k \right) \right) ^2$ ensures that force changes are provided as far as possible by the deformation of the elastic rope caused by the relative movement of human and robot, minimizing motor rotation.
    
    In the constraints, the human motion model is used to compute the system state. 
    In addition, the upper and lower bounds of force and the length, 
    the change of the force direction, the angle between yaw angle and the 
    force direction are constrained in \eqref{F}, \eqref{l}, \eqref{F direction}, 
    \eqref{F-yaw}, respectively. 
    We use one circle to cover the human and two circles to cover the robot as shown in Fig. \ref{fig:system formulation}
    to avoid obstacles by \eqref{obs_h_1} and \eqref{obs_d_1}. 
    
    \emph{2) Robot Motion Planning}: 
    The robot motion planning aims to obtain robot high-level velocity input. 
    The optimal reference $\left\{\boldsymbol{F}_{k}^{*},l_{k}^{*},\theta _{k}^{*} \right\} $ obtained from the previous stage is applied to robot motion planning. 
    Knowing $\boldsymbol{F}_{k}^{*}$ and the current state of the human, we can obtain the expected position and the human state $\left( \mathbf{x}_{k}^{h*},\mathbf{q}_{k}^{h*} \right)$ by \eqref{equ:qk+1} and \eqref{equ:xhk+1}. $\mathbf{x}_{k}^{d*}$ can be computed by \eqref{equ:dog_position} using $\left( l_{k}^{*},\theta _{k}^{*} \right)$. Moreover, robot position can be calculated by 

$
    \mathbf{\tilde{x}}_{k+1}^{d}=\mathbf{\tilde{x}}_{k}^{d}+\boldsymbol{DR}_{z,k}\mathbf{u}_kT
$
    where $\mathbf{\tilde{x}}^d=\left( x^d,y^d,\theta \right) ^{\top}$,
	and $\mathbf{u}=\left( \mathrm{v}_x,\mathrm{v}_y,\omega \right) ^{\top}$ represents robot's control input.
	$\boldsymbol{D}\in \mathbb{R} ^{3\times 3}$ is the velocity discount coefficient matrix caused by being pulled (identity matrix when in standing state), 
    $\boldsymbol{R}_z\in \mathbb{R} ^{3\times 3}$ is the rotation matrix. 
    Based on an MPC problem with a horizon of $M$ steps, we minimize the cost function
    \begin{equation}
    J\left( \mathbf{u} \right) =\sum_{k=0}^{M-1}{\left( \left\| \mathbf{\tilde{x}}_{k}^{d}-\mathbf{\tilde{x}}_{k}^{d*} \right\| _{\mathbf{Q}^d}^{2}+\left\| \mathbf{u}_k \right\| _{\mathbf{R}^d}^{2} \right)},
     \end{equation}
    where $\mathbf{Q}^d, \mathbf{R}^d \in \mathbb{R} ^{3\times 3}$ are positive definite. 
    The constraints \eqref{obs_h_1} and \eqref{obs_d_1} are used for obstacle avoidance.
    We formulate these two motion planning problems in CasADi \cite{andersson2019casadi} and solve them by IPOPT \cite{2009Large}.
    
\subsection{State Observation and Estimation}
\label{subsec:UKF}
Cartographer\cite{hess2016cartographer} is used for real-time mapping and robot localization, and DNN-based Face Detection and Recognition \cite{bradski2008learning} is used to get the human position. 
The camera is mounted on a gimbal with coordinates $\mathbf{x}^d$, providing rotation to maintain the face at the center of the image. We denote the average depth of facial feature points as $D_f$, gimbal's yaw as $\phi$, and the horizontal distance between human and camera as $l_c=D_f/\cos \varphi -\left( H-h_{\mathrm{c}} \right) \tan \varphi$, 
where $H$ is human's height and $\varphi$, $h_{\mathrm{c}}$ are the camera's fixed inclination angle and height, respectively. 
Then human position
$
    \mathbf{x}^h=\mathbf{x}^d-l_c\mathbf{e}_{\theta +\phi}
$ can be obtained.

Unscented Kalman Filter (UKF) is used for state estimation. The weight of the center point is set to be $\omega _0=k/\left( n+k \right) $, and the weight of the remaining points is  $\omega _i=1/2(n+k)$. 
When the human is walking, the system state can be computed by
\begin{equation}
\label{sys}
\mathbf{x}^+=\mathbf{x}+\left( \left[ \begin{matrix}
	\alpha \boldsymbol{I}_2&		\mathbf{0}\\
	\mathbf{0}&		\boldsymbol{DR}_z\\
\end{matrix} \right] \left[ \begin{array}{c}
	\boldsymbol{F}\\
	\mathbf{u}\\
\end{array} \right] +\left[ \begin{array}{c}
	\beta \mathbf{e}_F\\
	\mathbf{0}\\
\end{array} \right] \right) T+\mathbf{w}
\end{equation}
and when human is standing, the state can be computed by
\begin{equation}
\label{sys1}
\mathbf{x}^+=\mathbf{x}+\left( 0,\boldsymbol{R}_z\mathbf{u}T \right) ^{\top}+\mathbf{w}
\end{equation}
where 
$\mathbf{w}$ is process noise. The covariance of $\mathbf{w}$ is
$
\boldsymbol{Q}=\mathrm{diag}\left( \sigma _{F,x^h}^{2},\sigma _{F,y^h}^{2},\sigma _{x}^{2},\sigma _{y}^{2},\sigma _{\theta}^{2}\right)
$
, where
$\sigma _{x}^{2}$, $\sigma _{y}^{2}$, $\sigma _{\theta}^{2}$ are the variances of $x$, $y$, $\theta$, respectively, generated by the uncertainty in the motion of the quadruped robot, and $\sigma _{F,x^h}^{2}$, $\sigma _{F,y^h}^{2}$ are the error, generated by fitting $F-\mathrm{v}^{hl}$ relationship.
For each measurement, observation equation is as follows: 
\vspace{-0.2cm}
\begin{equation}
\label{ob}
\mathbf{\hat{z}}=\left( \hat{l},\hat{\phi},\hat{x}^d,\hat{y}^d,\hat{\theta} \right) ^{\top}=\left( l,\theta ^h-\theta ,x^d,y^d,\theta \right) ^{\top}+\mathbf{v}
\vspace{-0.2cm}
\end{equation}
where $\mathbf{v}$ is is process noise. And its covariance is $\boldsymbol{R}=\mathrm{diag}\left( \sigma _{l}^{2},\sigma _{\phi}^{2}, \sigma _{s,x}^{2},\sigma _{s,y}^{2},\sigma _{s,\theta}^{2} \right) $, introduced by human position measurement and Cartographer, respectively.

\subsection{Path Planning}
\label{subsec:Path Planning}
We use a heap-based A* planner over the grid map to generate a collision-free set of human waypoints. The human coordinate is extended to
$\mathbf{\tilde{x}}^h=\left( x^h,y^h,\theta ^h \right) ^{\top}\in \mathbb{R} ^3$.
The transition between nodes representing one step of a human is defined as $\left(\Delta L, \Delta\theta^h\right)$,
where $\Delta L=\sqrt{(\Delta x^h)^2+(\Delta y^h)^2}$ represents the human’s step size and $\Delta\theta^h$ represents the difference in the direction angle of the step. By limiting the size of $\Delta\theta^h$, the direction of human movement can not be greatly changed. The node cost $
g(\mathbf{\tilde{x}}_{n}^{h})=\sum_{i=1}^n{\left\| \mathbf{\tilde{x}}_{i}^{h}-\mathbf{\tilde{x}}_{i-1}^{h} \right\| _2}
$ and the heuristic cost $
h(\mathbf{\tilde{x}}_{n}^{h})=\left\| \mathbf{\tilde{x}}_{n}^{h}-\mathbf{\tilde{x}}_{\mathrm{target}}^{h} \right\| _2
$. We assume that human always move in the direction of force. Since $\mathbf {e}_F=\mathbf{e}_l$, robot's position can be calculated from $\theta^h$ and the rope's length. We use breadth-first search to determine whether there is a continuous collision-free trajectory for the robot to go from $\mathbf{x}_{n-1}^{d}$ to $\mathbf{x}_{n}^{d}$, in order to judge whether the human can transfer from $\mathbf{\tilde{x}}_{n-1}^{h}$ to $\mathbf{\tilde{x}}_{n}^{h}$.
\section{Experiments}
\subsection{Relationship Between Traction Force and Human Velocity}
\label{subsec:system_identification}
To develop the human motion model, we studied the relationship between human walking velocity and force. 
Measurements from different subjects reveal that the velocity of a human moving in the direction of the rope is roughly linearly related to the traction force and can be described as a linear function of slope $\alpha$ and intercept $\beta$. 

\begin{figure}[!htbp]
    \vspace{-0.3cm}
    \centering
    \includegraphics[width=8cm]{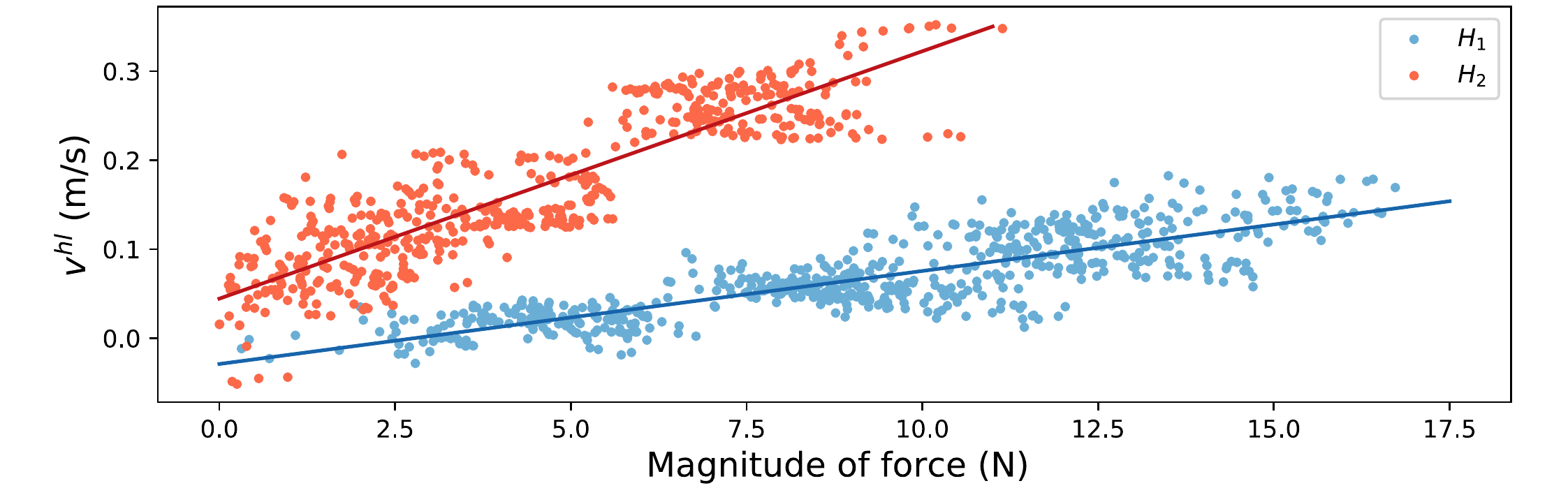}
    \vspace{-0.1cm}
    \caption{
    $F-\mathrm{v}^{hl}$ curves of two test subjects $H_1$, $H_2$. The parameters of the least-squares fit are:  $\alpha_1 = 0.0105$, $\beta_1 = -0.0290$; $\alpha_2 = 0.0278 $, $\beta_2 = 0.0444 $. 
     }
    \label{fig:adaptation}
    \vspace{-0.4cm}
\end{figure}

Fig. \ref{fig:adaptation} shows the different responses to force and the parameters of adaptation for two different subjects.

\subsection{Robot Guiding Human Experiments}

\begin{figure*}[ht]
\centering
\subfloat[]{
    \begin{minipage}[t]{0.21\linewidth}
        \centering
        \includegraphics[height=4.1cm]{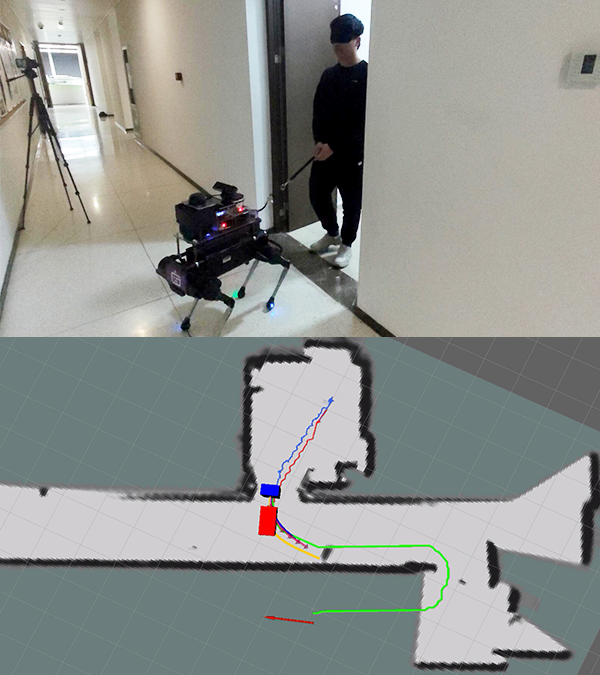}
        \label{fig:a}
        \vspace{-0.4cm}
    \end{minipage}
}
\subfloat[]{
    \begin{minipage}[t]{0.21\linewidth}
        \centering
        \includegraphics[height=4.1cm]{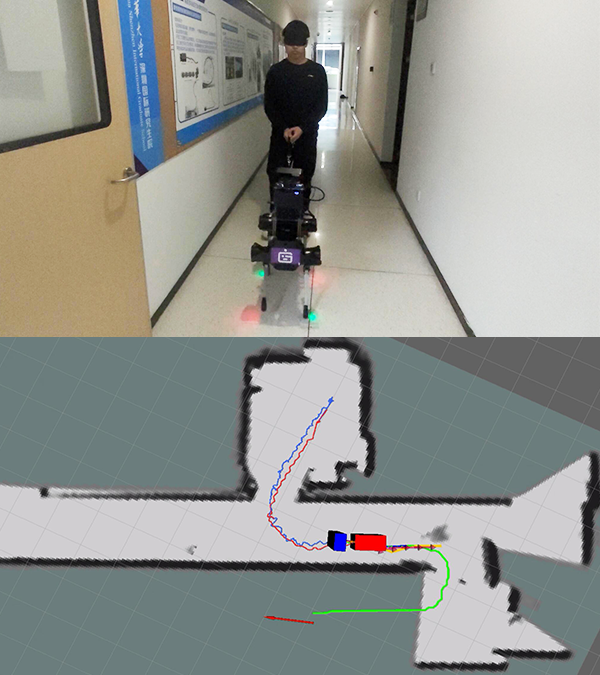}
        \label{fig:b}
        \vspace{-0.4cm}
    \end{minipage}
}
\subfloat[]{
    \begin{minipage}[t]{0.21\linewidth}
        \centering
        \includegraphics[height=4.1cm]{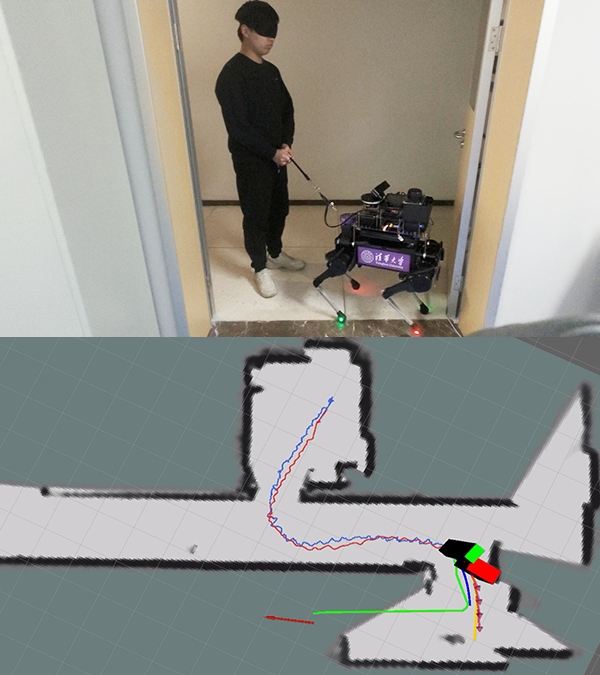}
        \label{fig:c}
        \vspace{-0.4cm}
    \end{minipage}
}
\subfloat[]{
    \begin{minipage}[t]{0.21\linewidth}
        \centering
        \includegraphics[height=4.1cm]{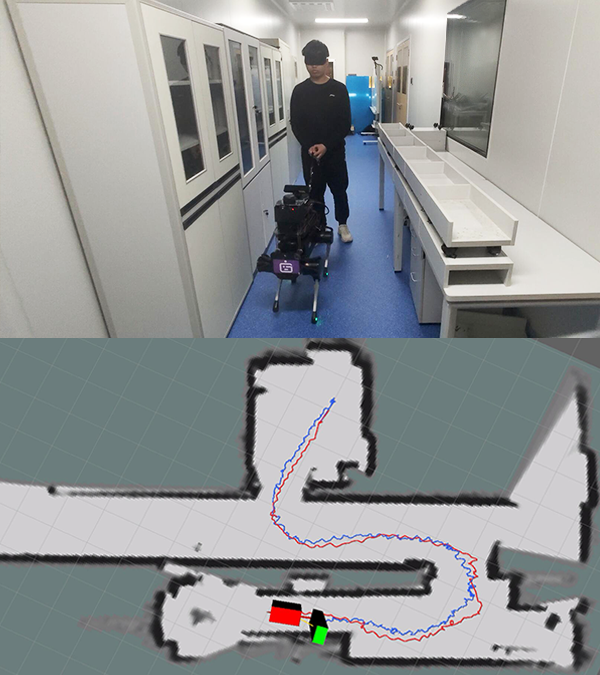}
        \label{fig:d}
        \vspace{-0.4cm}
    \end{minipage}
}
\subfloat{
\begin{minipage}[t]{0.1\linewidth}
    \centering
    \includegraphics[height=4.18cm]{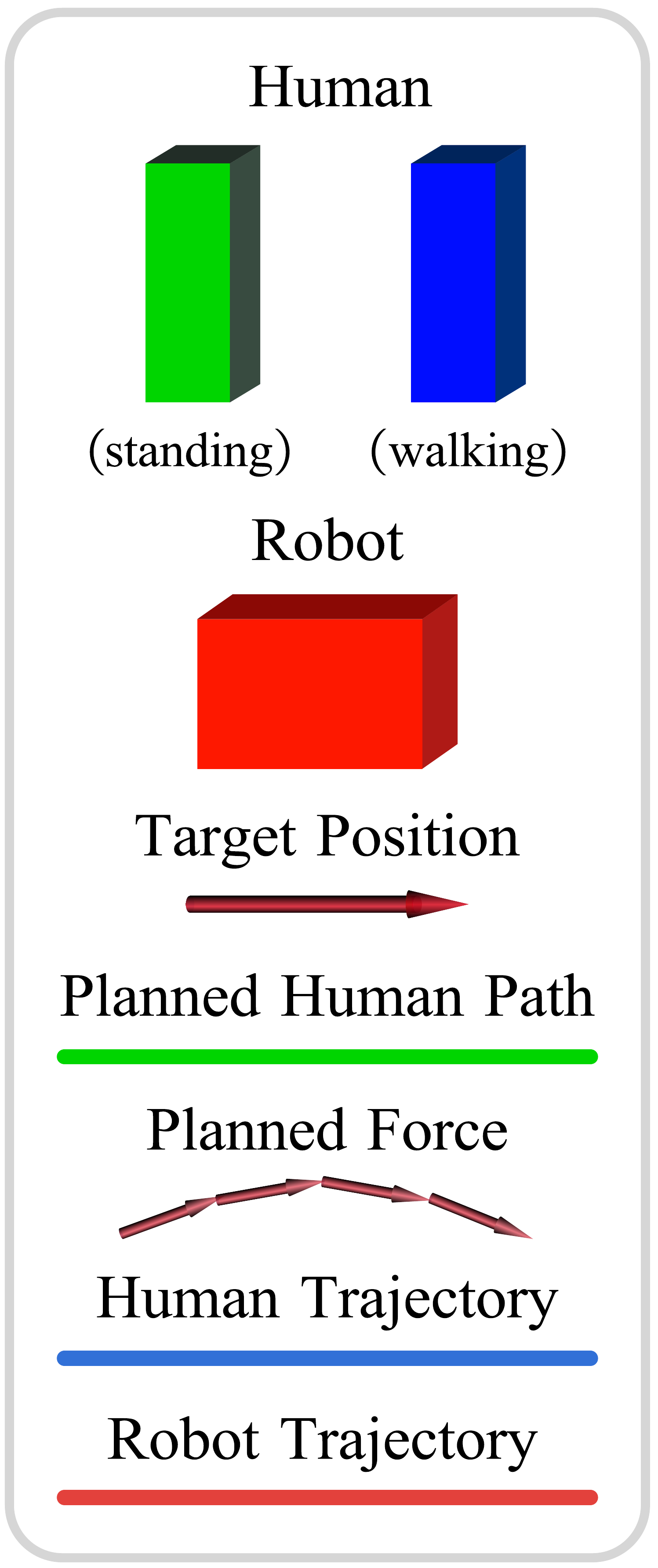}
    \vspace{-0.3cm}
\end{minipage}
}
\vspace{-0.1cm}
\caption{
    Snapshots of the guiding experiment in a real-world scenario. Laikago guided a blind-folded person from the elevator lobby, through two doors and two corridors, to the target position in the laboratory. The door closer to the starting point is 1.1m in width while the other is 1.3m in width, and the two corridor widths are 1.5m and 1m.
}
\vspace{-0.5cm}
\label{fig:exp}
\end{figure*}

\begin{figure}[!htbp]
    \centering
        \subfloat[Human trajectory with uncertainty and planned forces. ]{
            \begin{minipage}[t]{0.45\linewidth}
                \centering
                \includegraphics[width=3.8cm]{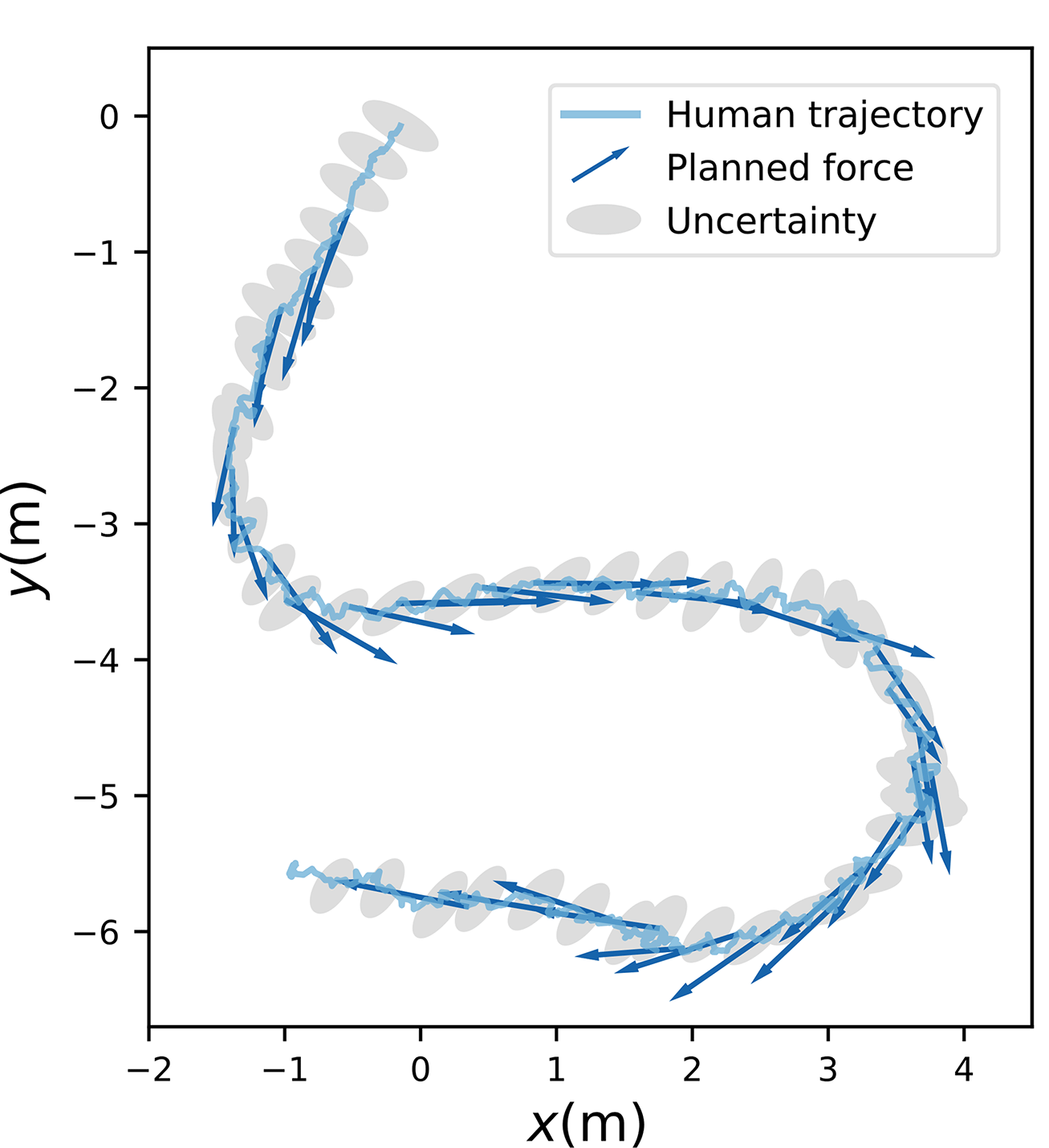}
            \end{minipage}
            \label{fig:human_trajectory}
        }
    \hfill
    \subfloat[Curves of the force magnitude for different methods at the same turn. ]{
            \begin{minipage}[t]{0.474782\linewidth}
                \centering
                \includegraphics[width=4.00927cm]{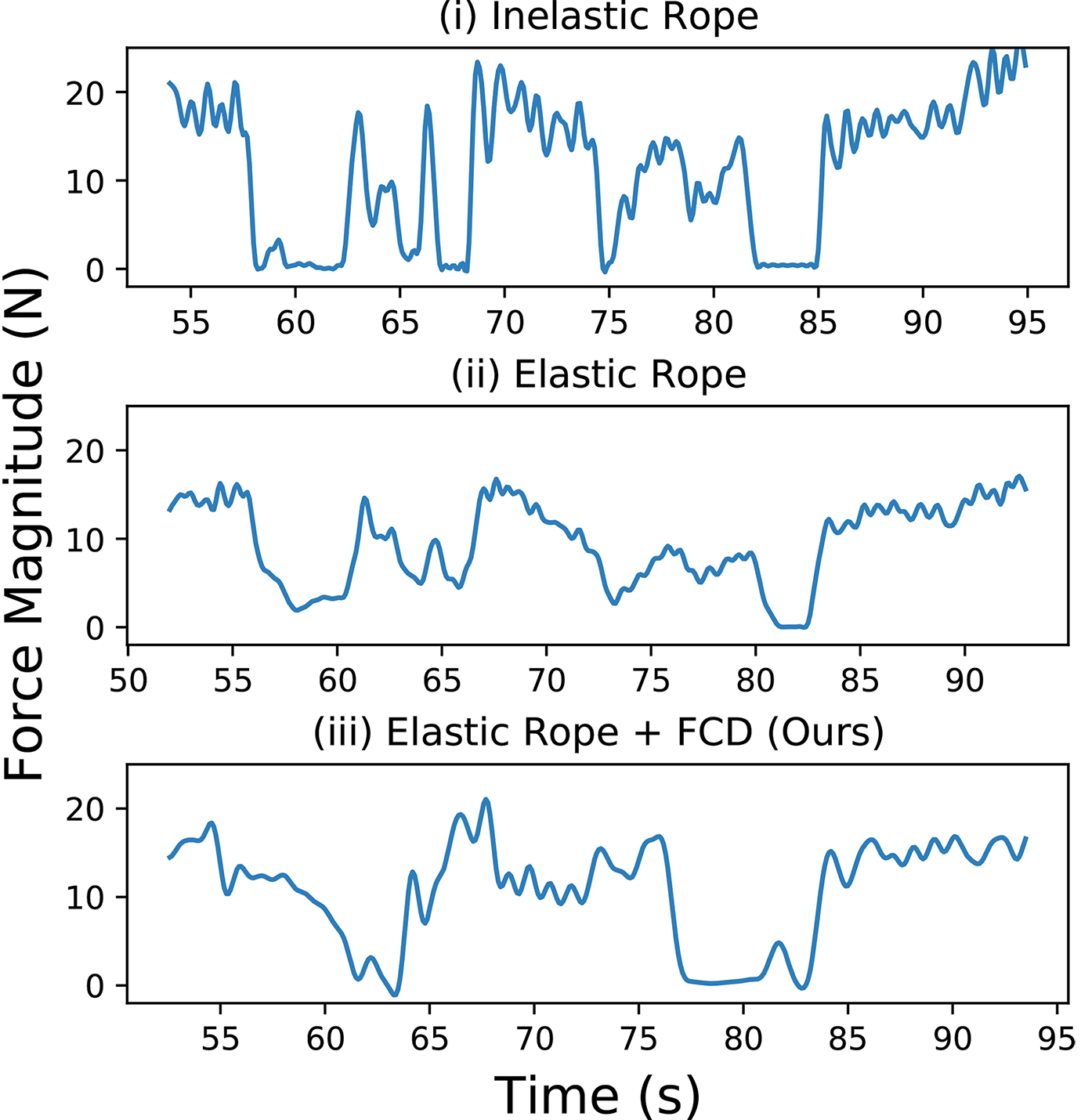}
            \end{minipage}
            \label{fig:force_compare}
        }
    \vspace{-0.2cm}
    \vfill
        \subfloat[Force magnitude and direction (the initial angle is set to 0). ]{
            \begin{minipage}[t]{0.970\linewidth}
                \centering
                \includegraphics[width=8.32cm]{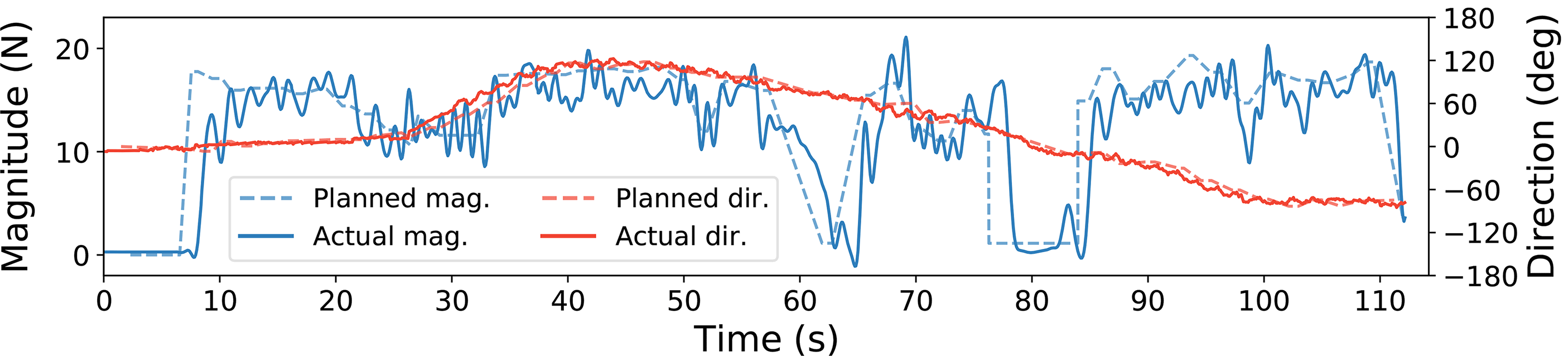}
            \end{minipage}
            \label{fig:force_result}
        }
    \caption{Experimental data of the guiding process. Fig. \ref{fig:human_trajectory} is the intercepted portion of the human trajectory (with 95\% confidence interval) and the planned force. Fig. \ref{fig:force_compare} depicts the force magnitude for the comparison experiment in the identical bend. Fig. \ref{fig:force_result} shows the magnitude and direction of the planned and actual force.}
    \label{fig:experiment_result}
    \vspace{-0.7cm}
\end{figure}

We conducted experiments in several real-world scenarios as shown in Fig. \ref{fig:exp}. 
Using UKF on the observations, we represented the uncertainty in the human's location as a gray ellipse in Fig. \ref{fig:human_trajectory}. 
As the robot began to move, the observation data was continuously updated, leading to a gradual decrease and stabilization of uncertainty for both the human and robot positions.

Due to the long length of the system configuration, accidents may occur if a rigid arm or an inelastic rope is used for guidance.
For example, the robot may repeatedly adjust and try to pull the human through, which will cause the person to repeatedly walk and stop many times in a short period and increase anxiety leading to discomfort.
In contrast, our approach can guide the human with more appropriate force and fewer pauses. 
During 61$\sim$64s and 76$\sim$83s shown in Fig. \ref{fig:force_result}, it was difficult for the human and robot to move and pass through the narrow bend at the same time. 
The human motion planner therefore caused the force to drop, and the human stopped after feeling it. 
At this point, the robot was free to move and the motor released the string to reduce the force ensuring that the human stands still without taking a step. 
After the robot had been adjusted to a position suitable for traction, the motor retracted the string and the elastic rope started to pull the human along. It can be seen that the human did not walk and stop due to repeated changes in force, resulting in uncertainty and discomfort.

We evaluate the comfort of different guiding approaches in the identical scenario using (\romannumeral1) inelastic rope (without FCD, e.g., \cite{xiao2021robotic}), 
(\romannumeral2) elastic rope (without FCD), 
and (\romannumeral3) elastic rope with FCD (our approach) for experiments.
To assess comfort, we define four metrics and calculate the relative comfort index (RCI) using TOPSIS 
\cite{hwang1981methods} shown in Table \ref{table:result}. $\dot{F}_{\mathrm{rms}}$ and $\dot{\theta}_{\mathrm{rms}}^{h}$ reflect human linear and angular acceleration, respectively, which are proxies for motion fluency and contact smoothness; $t_{F>F_{\max}}$ is the duration of force exceeding the specified maximum value, characterizing the appropriate traction force and walking velocity, with smaller values relieving fear; $N_{\mathrm{ch.}}$ is the number of state changes, fewer state transitions can reduce anxiety, ensure the fluency and predictability of motion. 
RCI is a weighted evaluation of the above four metrics. 

\begin{table}[!htbp]
    \caption{Results of Comfort Metrics for Different Approaches}
    \label{table:result}
    \resizebox{\linewidth}{!}{ 
    \begin{tabular}{|c|c|c|c|c|c|}
        \hline
        Methods & 
        $\dot{F}_{\mathrm{rms}}$ & 
        $\dot{\theta}_{\mathrm{rms}}^{h}$ &
        $t_{F>F_{\max}}$ &
        $N_{\mathrm{ch.}}$ & 
        RCI \\ 
        \hline 
        Inelas. & $16.700$ & $0.110$ & $17.166$ & $20$ & $0.1082$ \\ 
        \hline
        Elas. & $7.187$ & $0.117$ & $3.478$ & $14$ & $0.5905$ \\ 
        \hline
        Elas.+FCD (Ours) & $6.390$ & $0.104$ & $0.483$ & $6$ & $0.8820$ \\ 
        \hline \hline
        Planned & $5.681$ & $0.097$ & $0$ & $6$ & $1$ \\ 
        \hline
    \end{tabular}
    }
    \vspace{-0.4cm}
\end{table}
Table \ref{table:result} demonstrates that the comfort-based planning approach results in more appropriate force and reduces the number of state changes, outperforming (\romannumeral1) in all metrics. 
So \textit{Planned} has the best RCI, while (\romannumeral1) exhibits the worst. 
By comparing (\romannumeral1) and (\romannumeral2), it can be found that the elastic rope can act as a shock absorber against the change in force magnitude, thus significantly reducing $\dot{F}_{\mathrm{rms}}$, $t_{F>F_{\max}}$, $N_{\mathrm{ch.}}$ and improving RCI. 
With the addition of FCD, the traction force on the elastic rope can be controlled. Therefore the force of (\romannumeral3) is closer to the planned value and the metrics and RCI of (\romannumeral3) are further improved compared to (\romannumeral2). 
These conclusions can also be seen qualitatively in Fig.\ref{fig:force_compare}.
\section{Conclusion and Future Work}
\label{sec:conclusion}

This work proposes a comfort-based quadruped guidance robot system for planning and controlling traction for the visually impaired.
A force-based human motion model is developed to solve the appropriate magnitude and direction of traction forces in a human motion planner.
A traction device, including a force control device and an elastic rope, is designed to ensure force magnitude.
A robot motion planner is used to solve the robot velocity input, which ensures the force direction. 
The proposed approach was deployed and validated on Unitree Laikago to guide blind-folded people in narrow corridors.
Experiments comparing different guidance approaches supported the effectiveness of our system in improving comfort.
Future work will focus on more complex applications in large-scale scenarios, such as cross-floor guidance tasks and safe sidewalk navigation considering dynamic obstacles.


\newpage
\bibliographystyle{IEEEtran}
\bibliography{IEEEabrv, bib/bibliography}
\end{document}